%% file: main.tex
\documentclass[runningheads]{llncs}
\usepackage[T1]{fontenc}
\usepackage{graphicx}
\usepackage{xurl}
\usepackage{hyperref}
\usepackage{orcidlink}

\usepackage{amssymb}
\usepackage{booktabs}
\usepackage{subcaption}

\usepackage[acronyms]{glossaries}

\makeglossaries
\loadglsentries{acronyms.tex}

\newcommand{\urlGitHub}{https://github.com/bckrlab/cytobert}

\newcommand{\urlBckrLab}{https://bckrlab.org/p/cytobert}

\newcommand{\modelname}{CytoBERT}

\begin{document}
\title{\modelname{}: A Foundation Model for Single-Cell Cytometry~Data}
%
%
\author{
    Syed Abdul Haseeb Qadri\orcidlink{0009-0009-6471-5850}\inst{1}
    \and
    Bjarne C. Hiller\orcidlink{0009-0005-9371-1702}\inst{1}
    \and
    Felix Blanke\orcidlink{0009-0008-8950-0541}\inst{2}
    \and
    Vanja Sophie Cangalovic\orcidlink{0000-0002-0636-7829}\inst{2}
    \and
    Kutalmış Coşkun\orcidlink{0000-0001-7680-3182}\inst{2}
    \and
    Amin Mirzaei\orcidlink{0009-0001-5210-7956}\inst{2}
    \and
    Tom Siegl\orcidlink{0000-0003-2292-1188}\inst{1}
    \and
    Sebastian Bader\inst{1}
    \and
    Thomas Kirste\inst{1}
    \and
    Martin Becker\orcidlink{0000-0003-4296-3481}\inst{1,2,3}
}

\authorrunning{S.~A.~H.~Qadri et al.}

\institute{
    University of Rostock, Rostock, Germany
    \and
    Marburg University, Marburg, Germany
    \and
    Hessian Center for Artificial Intelligence, Darmstadt, Germany
}

\maketitle              
%
\input{sections/abstract}
\input{sections/01_introduction}
\input{sections/02_related_work}
\input{sections/03_methods}
\input{sections/04_results}
\input{sections/05_conclusion}

\input{sections/acknowledgements}

\newpage
%
%
%
%
\bibliographystyle{splncs04}
\bibliography{references}






\newpage

\appendix
\input{sections/appendix}

\end{document}

%% file: acronyms.tex
\newacronym{cytof}{CyTOF}{Cytometry by Time of Flight}

%% file: sections/abstract.tex
\begin{abstract}
Cytometry measures the complex characteristics of single cells (e.g., counts and protein expression of immune cells) and is widely used across immunological research and clinical settings.
However, cytometry data is highly heterogeneous and unstandardized due to experimental protocols and the choice of measured features.
While machine learning methods hold the potential to gain deeper insights into cell biology, these challenges make them difficult to apply and transfer across studies.
Recent advances in foundation models can alleviate these issues, but corresponding approaches are still scarce in this field.
To address this, we provide \modelname{}, a publicly available, open-source, open-weight foundation model for single-cell cytometry data with variable marker panels. \modelname{} is pretrained in a self-supervised manner on a large-scale cytometry corpus (15 human datasets with heterogeneous marker panels and more than 50 million cells) curated through marker standardization, enabling it to learn transferable inter-marker relationships within cells. Fine-tuning \modelname{} for sample-level classification demonstrates that transfer learning across heterogeneous cytometry datasets is feasible, providing a starting point for scalable, generalizable cytometry analysis.
Code is available on \href{\urlGitHub}{GitHub}.

\keywords{cytometry \and foundation models \and self‑supervised learning 
}
\end{abstract}

%% file: sections/01_introduction.tex
\section{Introduction}

Single-cell cytometry is widely used in immunology and clinical research for immunophenotyping, disease detection, and treatment-response analysis~\cite{hartmann2019comprehensive,robinson2023flow}. It measures the expression of 5--50 protein markers at single-cell resolution. Thus, each cytometry sample can be viewed as a cell-by-marker matrix, where rows correspond to individual cells and columns correspond to measured markers. Despite its broad applications, cytometry data remain difficult to integrate across studies~\cite{pedersen2022cycombine}. Depending on the study objective and laboratory, different sets of markers, known as marker panels, are measured~\cite{hu2022application}. In addition, even the same biological
marker may be associated with different detection channels, metal isotope tags, and
naming conventions~\cite{liu2020recent,overton2019reporting,hu2022application}. As a result, marker panels and marker annotations are generally not standardized across datasets, making it challenging to train models that can transfer knowledge across studies~\cite{hu2022application}.

Traditional cytometry analysis is commonly performed using manual gating, where bounding boxes are drawn around cell populations in a hierarchy of bivariate dot plots~\cite{liu2020recent}. Gated populations are then summarized per sample using population abundances and marker-expression intensities, which can be statistically associated with phenotypes or compared between clinical groups~\cite{nowicka2017cytof}. Although widely used, manual gating is laborious, subjective, time-consuming, and difficult to scale as the number of measured markers increases. Machine-learning and deep-learning methods have been proposed to automate cytometry analysis, but most are dataset-specific, assume a fixed marker panel, require labeled training data, and must be retrained for each new study or task~\cite{hu2022application,ding2025immunefm}.

To overcome these limitations, we introduce \modelname{}, a Transformer-based foundation model for cytometry data inspired by the recently proposed ImmuneFM framework~\cite{ding2025immunefm}.
Unlike ImmuneFM, we provide implementation details and make code and data available. \modelname{} processes each cell independently as a sequence of marker-expression tokens and is pretrained with a self-supervised masked expression prediction objective to learn marker co-expression patterns within cells. We extend this approach with a shared vocabulary of 220 distinct markers and evaluate the pretrained model for transfer learning across heterogeneous cytometry datasets. The model can then be fine-tuned for downstream tasks such as sample-level disease classification.

%% file: sections/02_related_work.tex
\section{Related Work}

\paragraph{Sample classification in cytometry.}
Several methods have been proposed for cytometry-based sample classification. Classical approaches such as CITRUS~\cite{bruggner2014automated}, FloReMi~\cite{vangassen2016floremi}, and CytoDx~\cite{hu2019robust} extract cluster- or marker-level features from single-cell data for sample-level prediction. Deep-learning methods, including CellCNN~\cite{arvaniti2017sensitive}, CytoSet~\cite{yi2021cytoset}, and CytoCoSet~\cite{chen2025conditional}, learn sample representations directly from single-cell data using convolutional or permutation-invariant architectures. However, these methods are typically supervised and dataset-specific, assume a fixed marker panel, and require retraining for each new dataset or task.

\paragraph{Foundation models for single-cell data.} Foundation models have gained attention in single-cell transcriptomics, e.g., scBERT~\cite{yang2022scbert}, scGPT~\cite{cui2024scgpt}, scFoundation~\cite{hao2024largescale}, Geneformer~\cite{theodoris2023transfer}, CellFM~\cite{zeng2025cellfm}, and GeneMamba~\cite{qi2025genemamba}, which learn reusable representations from large-scale RNA-seq datasets. However, these methods target transcriptomics, where genes provide a stable feature vocabulary across datasets. Cytometry is more heterogeneous because studies measure different subsets of protein markers and often lack standardized marker annotations. Existing cytometry foundation models remain limited: cyMAE~\cite{kim2024cytometry} assumes a fixed 30-marker panel (MDIPA) and is evaluated only on cohorts sharing that panel, while ImmuneFM~\cite{ding2025immunefm} addresses variable-panel cytometry but does not provide a publicly available implementation. \modelname{} addresses this gap by learning transferable marker-expression representations from standardized cytometry datasets with variable marker panels and evaluating them for downstream sample-level classification.

%% file: sections/03_methods.tex
\section{Method}

The core of \modelname{} is a Transformer encoder parametrized by $\theta$ and pretrained using masked expression modeling. Each cell is represented as a sequence of tokens encoding marker--expression pairs. Let $m_j \in \{1,\ldots,M\}$ denote the marker index of token $j$, where $M=220$ is the number of distinct markers. \modelname{} learns a marker embedding function $E^{m}: \{1,\ldots,M\} \rightarrow \mathbb{R}^{d}$, with $d=128$. Expression values are transformed using $\mathrm{arsinh}(x/5)$, MinMax-scaled per marker across the pretraining corpus, and discretized into $B=10$ equal-width expression bins. Let $b_j \in \{1,\ldots,B\}$ denote the corresponding bin index. \modelname{} learns a bin embedding function $E^{b}: \{1,\ldots,B\} \rightarrow \mathbb{R}^{d}$. The final token embedding, $\mathbf{h}_j$, is given by
\[
\mathbf{h}_j = E^{m}(m_j) + E^{b}(b_j).
\]
Thus, each input token $\mathbf{h}_j$ combines marker identity and discretized expression information, enabling \modelname{} to process cells with variable marker panels.

\subsection{Self-Supervised Pretraining}
\label{sec:pretraining}

During pretraining, a subset of measured marker positions is randomly selected for masking. For this, the expression-bin embedding is replaced by a special \texttt{[MASK]} embedding, while the marker identity remains visible. This forces the model to infer the missing expression state of a known marker from the remaining marker-expression context within the same cell. Since zero-valued expressions dominate cytometry data and are often less informative, masking is applied only to non-zero marker-expression positions. A detailed overview of the framework is provided in supplementary material Fig.~\ref{fig:cytofm_pipeline}.

Following Transformer-based single-cell models such as scBERT~\cite{yang2022scbert} and scGPT~\cite{cui2024scgpt}, we prepend a learnable classification token to the input sequence. During pretraining, the masked-bin prediction objective is applied to marker-token representations, while the classification token is retained for downstream fine-tuning as a compact cell-level representation:
\[
\mathbf{H}_{in} = [\mathbf{h}_{\mathrm{CLS}}, \mathbf{h}_1, \dots, \mathbf{h}_L],
\quad
\mathbf{H} = \mathrm{Transformer}(\mathbf{H}_{in}).
\]
The output $\mathbf{H}_0$ corresponds to the classification token used during downstream fine-tuning, while $\mathbf{H}_j$ denotes the contextualized representation of marker position $j$ used for masked-bin prediction. A linear prediction head maps masked marker representations to logits over expression bins,
\[
\mathbf{z}_j = \mathbf{W}\mathbf{H}_j + \mathbf{b}, \qquad \mathbf{z}_j \in \mathbb{R}^{B}.
\]
The logits define a conditional probability distribution over expression bins at masked marker position $j$,


\[
p_{\theta}(B_j=b \mid m_j, \{m_i, b_i\}_{i \notin \Omega}) =
\mathrm{softmax}(\mathbf{z}_j)_b ,
\]

where the variable $B_j$ corresponds to the expression-bin label at position~$j$, and $\Omega$ is the set of masked positions. The pretraining objective minimizes cross-entropy over masked positions only:


\[
\mathcal{L}_{pretrain}
=
-\frac{1}{|\Omega|}
\sum_{j \in \Omega}
\log p_{\theta}(B_j=b_j \mid m_j, \{m_i, b_i\}_{i \notin \Omega}).
\]

By reconstructing masked bins from the remaining measured markers, \modelname{} learns contextual inter-marker relationships within individual cells.

\subsection{Fine-Tuning for Sample Classification}
\label{sec:finetuning}

After pretraining, \modelname{} is fine-tuned end-to-end for supervised sample-level classification. Since labels are defined at the sample or patient level, cell-level representations must be aggregated into a sample-level embedding. For a sample containing $K$ cells, each cell $k$ is processed independently by the pretrained encoder, and the classification-token output is used as the cell embedding: $\mathbf{c}_k = \mathbf{H}^{(k)}_0$.
Cell embeddings are then aggregated using mean pooling,
\[
\mathbf{s} = \frac{1}{K}\sum_{k=1}^{K}\mathbf{c}_k,
\]
where $\mathbf{s} \in \mathbb{R}^{d}$ is the sample representation. This representation is passed to a task-specific, two-layer MLP classification head. During fine-tuning, both the pretrained encoder and the classification head are optimized jointly using cross-entropy loss. In the main experiments, we sample $K=512$ cells per sample and evaluate performance using ROC-AUC.

%% file: sections/04_results.tex
\section{Results \&{} Analysis}

\subsection{Experimental Setup}

\subsubsection{Pretraining setup.}
\modelname{} is pretrained on a curated CyTOF corpus consisting of approximately 50 million cells from 15 out of 18 curated datasets. During pretraining, 25\% of eligible non-zero marker-expression bins are masked in each epoch. The Transformer encoder uses hidden dimension $d=128$, 6 layers, 8 attention heads, and 10 expression bins, and is optimized with Adam for 50 epochs. These hyperparameters are primarily motivated by ImmuneFM~\cite{ding2025immunefm}, the closest related cytometry foundation-model setting. A detailed overview of the markers across all datasets is provided in Supplementary Fig.~\ref{fig:marker_vocab}.

\subsubsection{Fine-tuning setup.}
For downstream evaluation, we fine-tune \modelname{} on sample classification tasks across seven CyTOF datasets from ImmPort. Since a single sample or file may contain a large number of cells, we randomly select 512 cells without replacement from each file to create a smaller subset/subsample. To reduce sampling variability, multiple independent subsamples are generated from each sample and classified independently. Class imbalance is handled by oversampling minority classes at the sample level.

\subsubsection{Evaluation.}
We compare pretrained \modelname{} with logistic regression, CytoSet, and a randomly initialized \modelname{} model without pretrained weights (from scratch). Logistic regression uses marker expression values averaged across all cells in each subsample. CytoSet and \modelname{} both use a mean-pooling layer for sample-level aggregation. CytoSet uses a supervised, permutation-invariant architecture. Since logistic regression and CytoSet require fixed input dimensionality, they use only markers shared across all samples. \modelname{} is evaluated using its native variable-marker representation. Models are evaluated with 3-fold cross-validation. When subject identifiers are available, we use grouped splits to prevent subsamples from the same subject from appearing in both training and test folds. Performance is reported using ROC-AUC, with macro-averaged one-vs-rest ROC-AUC for multi-class tasks.

\begin{table}[!htbp]
\centering
\renewcommand{\arraystretch}{1.1}
\setlength{\tabcolsep}{4pt}
\caption{Downstream CyTOF datasets. ``Subjects'' and ``Samples'' denote unique patients and sample files.}
\label{tab:cytometry_datasets_summary}
\begin{tabular}{lcccc}
\toprule
\textbf{Dataset} & \textbf{Subjects} & \textbf{Samples} & \textbf{Disease/condition} & \textbf{In pretrain?} \\
\midrule
SDY788  & 20  & 126 & Transplant & Yes \\
SDY997  & 54  & 259 & Autoimmune & No \\
SDY1535 & 50  & 100 & HIV diagnosis & Yes \\
SDY1708 & 74  & 146 & COVID severity & No \\
SDY1733 & 49  & 99  & Liver cancer stage & No \\
SDY2011 & 113 & 185 & COVID diagnosis & Yes \\
SDY2015 & 36  & 107 & Peanut allergy & Yes \\
\bottomrule
\end{tabular}
\end{table}

\subsection{Results on Downstream Datasets}

\begin{figure}[!htbp]
    \centering

    {\footnotesize\textbf{Grouped Subject Split}}

    \includegraphics[width=\textwidth]{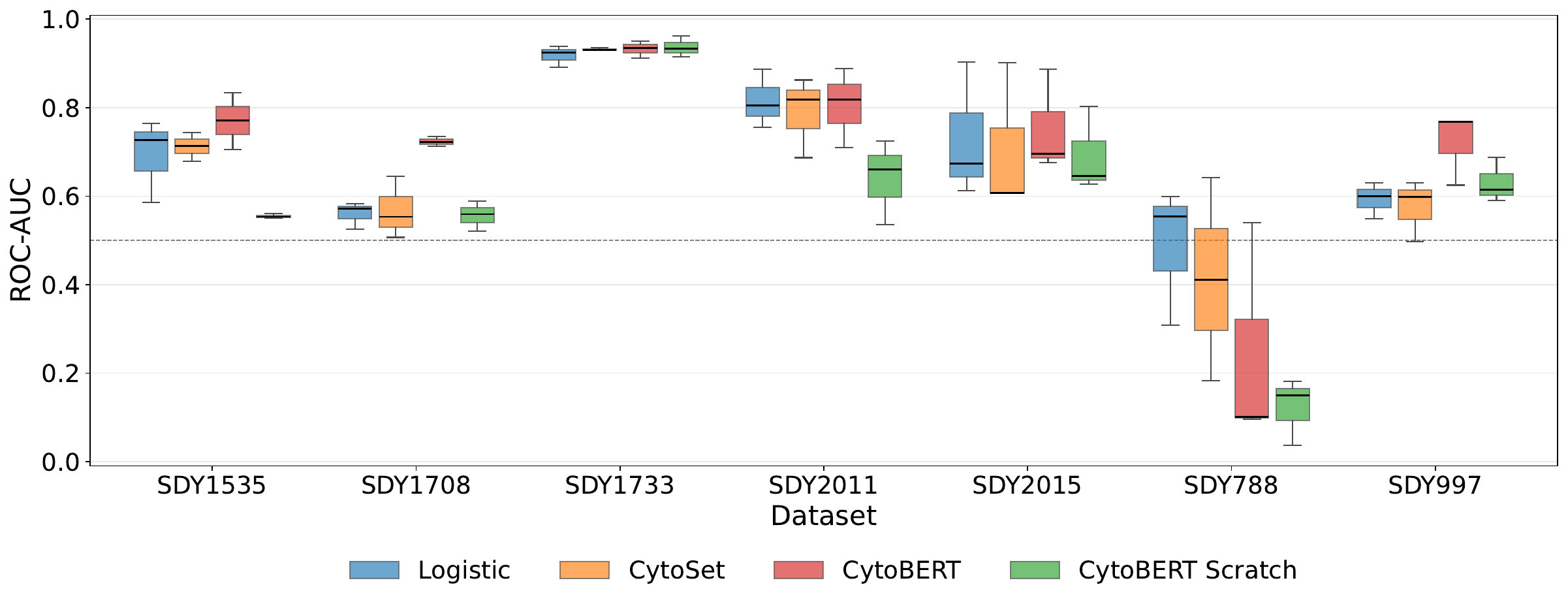}

    \caption{Sample classification results under grouped subject splitting. ROC-AUC is shown on the y-axis. The Friedman test found no significant differences between models ($p=0.093$), although \modelname{} achieved the best average rank.}
    \label{fig:auc_grouped}
\end{figure}

Sample-level prediction results are reported in Fig.~\ref{fig:auc_grouped}. Under grouped subject splitting, \modelname{} fine-tuning achieves the best average ranking across downstream datasets (Fig.~\ref{fig:cd_grouped_ungrouped}). Compared with \modelname{} trained from scratch, this suggests that self-supervised pretraining provides a useful initialization for sample classification. The greatest improvements are observed on SDY1708 and SDY997, indicating transfer to datasets not seen during pretraining.

We evaluate \modelname{} on some datasets included in pretraining (Tab.~\ref{tab:cytometry_datasets_summary}). However, pretraining uses only marker-expression values and no sample labels, avoiding label leakage. In addition, the clearest gains are observed on truly held-out datasets, such as SDY1708 and SDY997. Because SDY788 has few subjects, its grouped-split performance is weakest.

Unlike prior evaluations using ungrouped splits~\cite{yi2021cytoset,ding2025immunefm}, we report the main results under grouped subject splitting. This prevents samples from the same subject from appearing in both training and test folds, providing a more realistic estimate of generalization to unseen subjects. A comparison with ungrouped splitting is provided in Supplementary Figs.~\ref{fig:auc_grouped_ungrouped} and~\ref{fig:cd_grouped_ungrouped}.

%% file: sections/05_conclusion.tex
\section{Conclusion}

We presented \modelname{}, an open-source, open-weight foundation model for variable-panel single-cell cytometry data. By standardizing markers across cytometry datasets and self-supervised pretraining, \modelname{} can be fine-tuned for downstream sample classification. Across the evaluated datasets, \modelname{} achieves the best average ranking under grouped subject splitting, suggesting that pretraining improves sample-level classification performance. Although differences to supervised baselines are not statistically significant, the results indicate the potential of foundation-model-style pretraining for cytometry. Our code and trained weights are publicly available to support reproducibility and future work.


%% file: sections/acknowledgements.tex
\paragraph{Acknowledgments.} 
This work has been funded by the ERDF~(TBI-1-143-W-043), by the BMBF~(01IS22077, 01ED2507) and DFG~(CRC 1713: 534829736).


%% file: sections/appendix.tex
\section{Data and Code Availability}

A general overview of information about this manuscript is available at \url{\urlBckrLab}. 
Code is available at \url{\urlGitHub}. 
To ensure long-term reproducibility and accessibility of the results, 
we additionally provide a code snapshot \cite{qadri2026code}, 
data artifacts \cite{qadri2026data},
as well as a Docker image \cite{qadri2026docker}.

This publication uses data provided by
\href{https://www.immport.org}{ImmPort}.
The full dataset contains pre-processed data from the following ImmPort datasets:

\begin{itemize}
\item
  \href{https://www.immport.org/shared/study/SDY207/summary}{SDY207}:
  Cytometry by time-of-flight shows combinatorial cytokine expression
  and virus-specific cell niches within a continuum of CD8+ T cell
  phenotypes
\item
  \href{https://www.immport.org/shared/study/SDY461/summary}{SDY461}:
  Monitoring of tissue-specific immune responses in man to naturally
  occurring pathogens using mass cytometric monitoring
\item
  \href{https://www.immport.org/shared/study/SDY515/summary}{SDY515}:
  Monozygotic and Dizygotic Twin Pair T-Cell Responses to Influenza
  Vaccination SLVP018 2010
\item
  \href{https://www.immport.org/shared/study/SDY788/summary}{SDY788}:
  Immune Profiles to Predict Response to Desensitization Therapy in
  Highly HLA-Sensitized Kidney Transplant Candidates
\item
  \href{https://www.immport.org/shared/study/SDY997/summary}{SDY997}:
  AMP Lupus Network Project: Molecular Characterization of Lupus
  Nephritis and Correlation with Response to Therapy~
\item
  \href{https://www.immport.org/shared/study/SDY998/summary}{SDY998}:
  AMP Rheumatoid Arthritis Phase 1
\item
  \href{https://www.immport.org/shared/study/SDY1479/summary}{SDY1479}:
  Plasmablast response to inactivated and live attenuated influenza
  vaccines (TIV3/TIV3 ID) in 2014
\item
  \href{https://www.immport.org/shared/study/SDY1535/summary}{SDY1535}:
  TIGIT is upregulated by HIV-1 infection and marks a highly functional
  adaptive and mature subset of natural killer cells
\item
  \href{https://www.immport.org/shared/study/SDY1708/summary}{SDY1708}:
  Broad dysregulation of innate immunity and hematopoiesis distinguishes
  mild from severe COVID-19
\item
  \href{https://www.immport.org/shared/study/SDY1733/summary}{SDY1733}:
  Single-Cell immune signature for detecting early-stage HCC and early
  assessing PD-1 immunotherapy efficacy
\item
  \href{https://www.immport.org/shared/study/SDY1773/summary}{SDY1773}:
  Human plasmacytoid dendritic cells mount a distinct antiviral response
  to virus-infected cells
\item
  \href{https://www.immport.org/shared/study/SDY2011/summary}{SDY2011}:
  A shift in lung macrophage composition is associated with COVID-19
  severity and recovery
\item
  \href{https://www.immport.org/shared/study/SDY2015/summary}{SDY2015}:
  Mass cytometry analysis of PBMCs from peanut-sensitized tolerant and
  clinically allergic infants
\item
  \href{https://www.immport.org/shared/study/SDY2189/summary}{SDY2189}:
  Systems vaccinology of the BNT162b2 mRNA vaccine in humans
\item
  \href{https://www.immport.org/shared/study/SDY2471/summary}{SDY2471}:
  U19 CCHI\_Project 1. The Role of CD4 Memory Phenotype, Memory, and
  Effector T Cells in
\item
  \href{https://www.immport.org/shared/study/SDY2497/summary}{SDY2497}:
  Effects of Aging on Primary and Secondary Vaccine Responses in a
  15-Year Longitudinal Cohort
\item
  \href{https://www.immport.org/shared/study/SDY2739/summary}{SDY2739}:
  Clinical Immunity to Malaria Involves Epigenetic Reprogramming of
  Innate Immune Cells
\item
  \href{https://www.immport.org/shared/study/SDY2823/summary}{SDY2823}:
  The Sound Life Project Healthy Adult Immunotypes
\end{itemize}

\section{Results under Grouped and Ungrouped Splitting}

\begin{figure}[!htbp]
    \centering

    {\footnotesize\textbf{Ungrouped Subject Split}}

    \includegraphics[width=\textwidth]{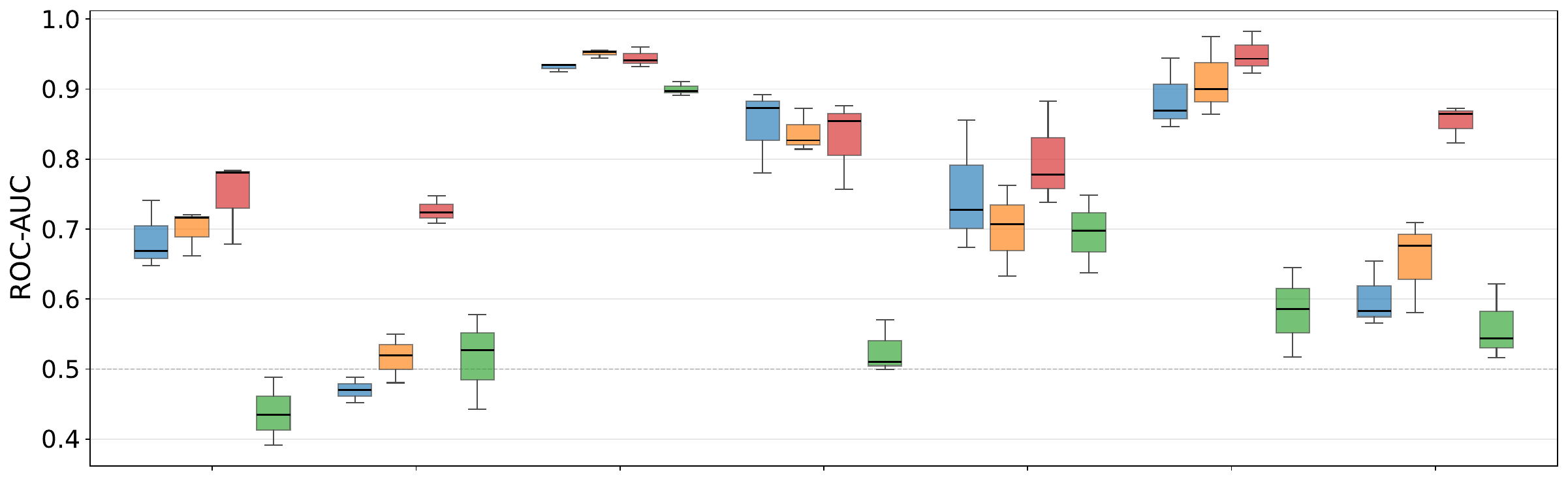}

    \vspace{0.5em}

    {\footnotesize\textbf{Grouped Subject Split}}

    \includegraphics[width=\textwidth]{figures/results/auc_boxplots_grouped.pdf}

    \caption{Sample-level classification performance across CyTOF datasets under ungrouped splitting (top) and grouped subject splitting (bottom). The y-axis reports ROC-AUC. Grouped subject splitting ensures that samples from the same subject do not appear in both training and test folds.}
    \label{fig:auc_grouped_ungrouped}
\end{figure}

A performance comparison in terms of ROC-AUC across models and datasets is shown in Fig.~\ref{fig:auc_grouped_ungrouped}. A substantial drop in performance is observed under grouped subject splitting, particularly for the SDY788 dataset. This suggests that ungrouped sample-level splitting may allow models to exploit subject-specific patterns rather than disease-specific signals, leading to overly optimistic performance estimates. Therefore, grouped subject splitting provides a more appropriate evaluation setting for sample-level classification.

\section{Comparison of Model Rankings}
On average, \modelname{} achieves the best rank under both splitting strategies, as shown in the critical difference diagrams below. Models connected by a horizontal line are not significantly different. Evaluation on additional datasets would provide a more robust comparison of model performance.

\begin{figure}[!htbp]
    \centering

    {\footnotesize\textbf{Ungrouped Subject Split}}

    \includegraphics[width=0.9\textwidth]{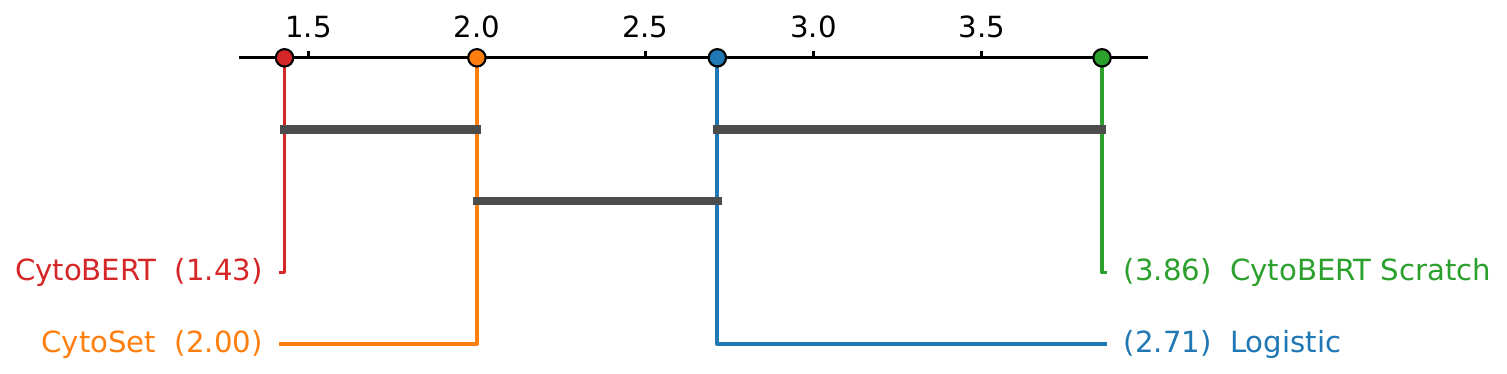}

    \vspace{0.5em}

    {\footnotesize\textbf{Grouped Subject Split}}

    \includegraphics[width=0.9\textwidth]{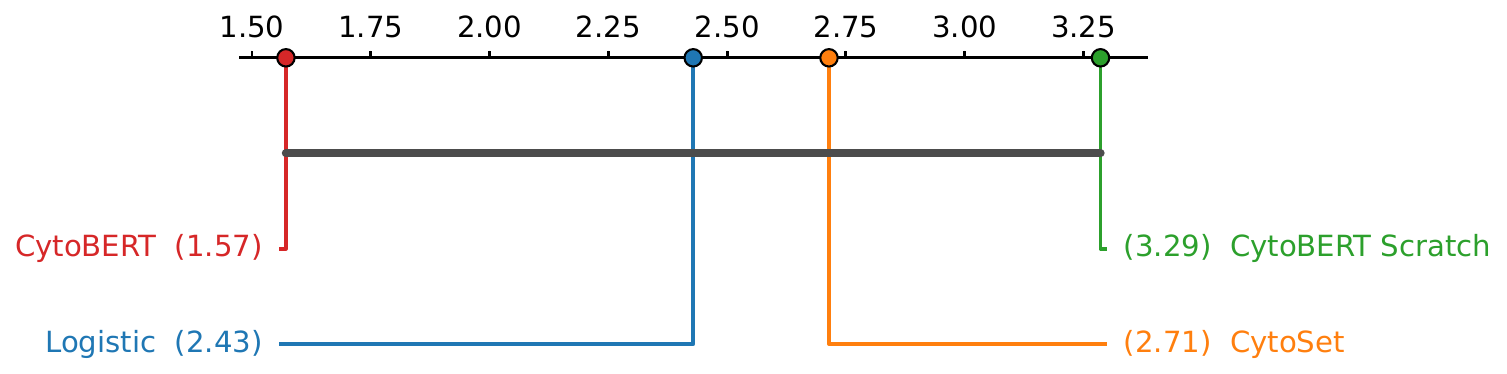}

    \caption{Critical difference diagrams comparing average model ranks across CyTOF datasets under ungrouped splitting (top) and grouped subject splitting (bottom). Lower rank indicates better average performance.}
    \label{fig:cd_grouped_ungrouped}
\end{figure}

\begin{table}[!htbp]
\centering
\caption{Mean ROC-AUC of different models across downstream datasets with grouped subject split.}
\label{tab:roc_auc_results}
\begin{tabular}{l@{\hspace{1.0cm}}c@{\hspace{0.8cm}}c@{\hspace{0.8cm}}c@{\hspace{0.8cm}}c}
\toprule
\textbf{Dataset} & \textbf{Logistic} & \textbf{CytoSet} & \textbf{\modelname{} Scratch} & \textbf{\modelname{}} \\
\midrule
SDY1708 & 0.5599 & 0.5685 & 0.5563 & \textbf{0.7234} \\
SDY1733 & 0.9180 & 0.9319 & \textbf{0.9366} & 0.9324 \\
SDY997  & 0.5929 & 0.5752 & 0.6310 & \textbf{0.7207} \\
SDY2015 & 0.7300 & 0.7050 & 0.6919 & \textbf{0.7526} \\
SDY1535 & 0.6922 & 0.7120 & 0.5549 & \textbf{0.7702} \\
SDY788  & \textbf{0.4875} & 0.4121 & 0.1229 & 0.2461 \\
SDY2011 & \textbf{0.8159} & 0.7891 & 0.6403 & 0.8050 \\
\bottomrule
\end{tabular}
\end{table}

\section{Overview of \modelname{}}
\begin{figure}[!htbp]
\centering

\begin{subfigure}{\linewidth}
    \centering
    \includegraphics[width=\linewidth]{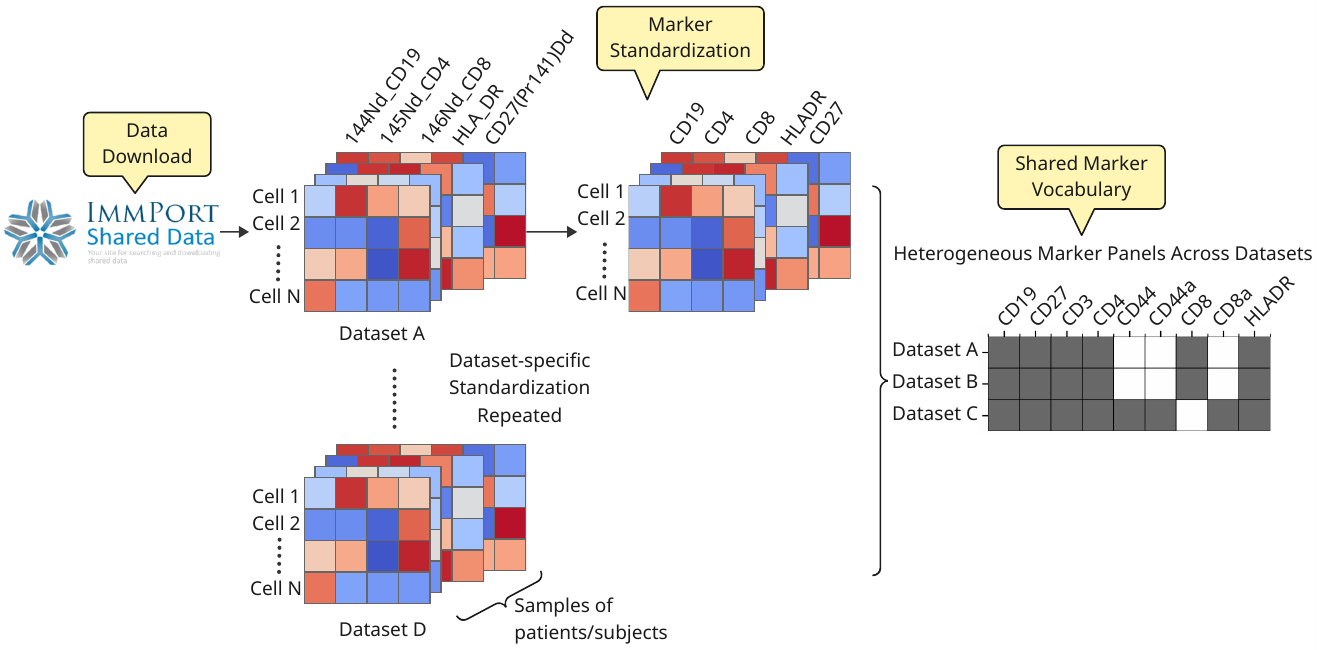}
    \caption{Dataset curation and preprocessing, including marker harmonization, transformation, scaling, and expression binning.}
    \label{fig:preprocess}
\end{subfigure}

\vspace{0.4em}

\begin{subfigure}{\linewidth}
    \centering
    \includegraphics[width=\linewidth]{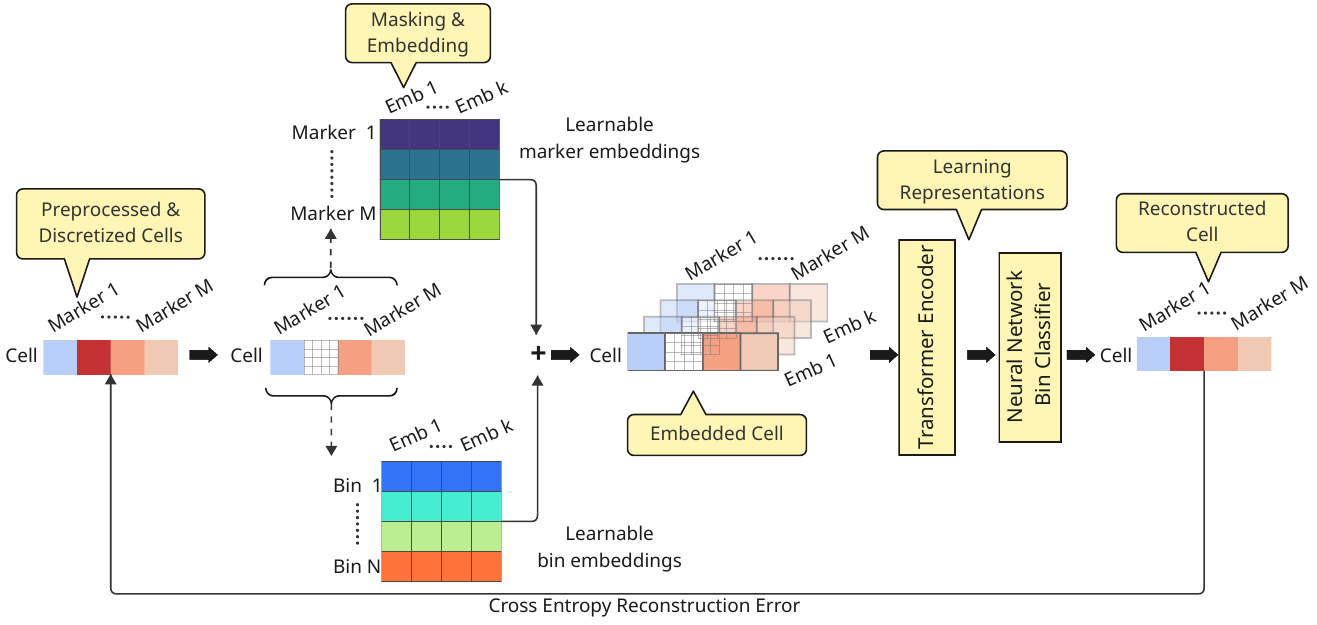}
    \caption{Self-supervised pretraining of \modelname{} for a single cell}
    \label{fig:pretrain}
\end{subfigure}

\vspace{0.4em}

\begin{subfigure}{\linewidth}
    \centering
    \includegraphics[width=\linewidth]{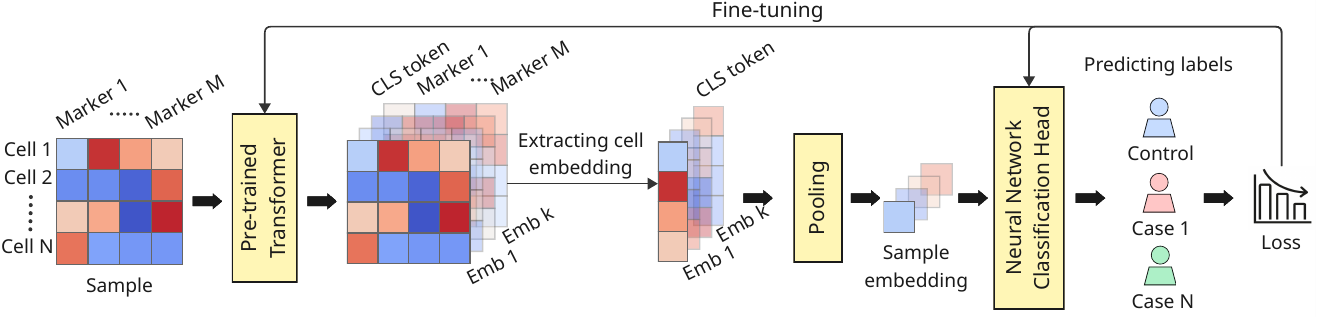}
    \caption{Fine-tuning the pretrained encoder for downstream sample-level classification.}
    \label{fig:finetune}
\end{subfigure}

\caption{Overview of the \modelname{} pipeline, consisting of dataset curation and preprocessing, self-supervised masked expression-bin pretraining, and supervised fine-tuning for sample-level classification.}
\label{fig:cytofm_pipeline}

\end{figure}

\begin{figure}[htbp]
\centering
\includegraphics[
    width=1\textwidth,
    height=1\textheight,
    keepaspectratio
]{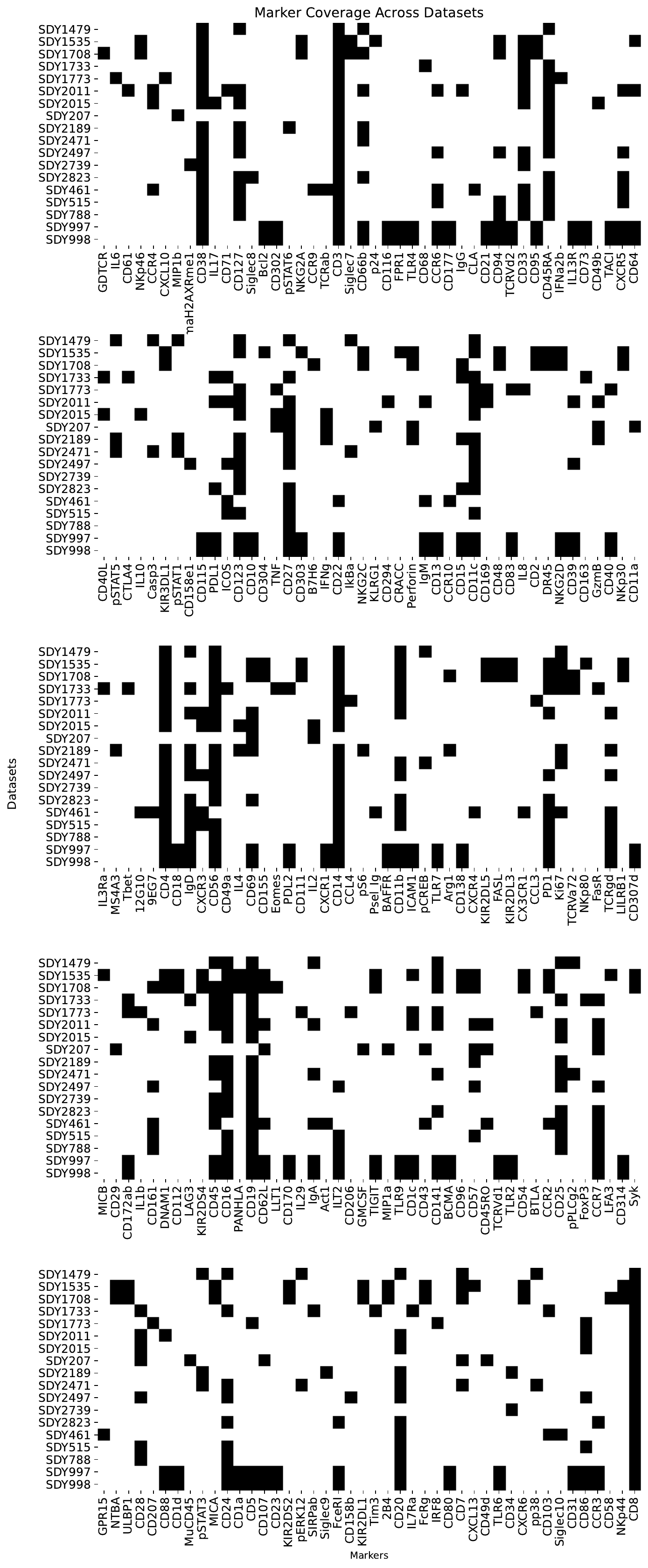}
\caption{Shared marker vocabulary across all CyTOF datasets used in this paper, consisting of 220 unique biomarkers in total. Black = marker measured/available.}
\label{fig:marker_vocab}
\end{figure}

Fig.~\ref{fig:cytofm_pipeline} shows an overview of the full \modelname{} pipeline, and Fig.~\ref{fig:marker_vocab} shows the shared marker vocabulary across all 18 datasets used.

